# Dimensionality Reduction and Dynamical Mode Recognition of Circular Arrays of Flame Oscillators Using Deep Neural Network

Weiming Xu[a,#], Tao Yang [b,#], and Peng Zhang[a,*]

*a Department of Mechanical Engineering, City University of Hong Kong, Kowloon Tong, Kowloon, 999077, Hong Kong*
*b Department of Mechanical Engineering, The Hong Kong Polytechnic University, Hung Hong, Kowloon, 999077, Hong Kong*

**Abstract**

Oscillatory combustion in aero engines and modern gas turbines often has significant adverse effects on their operation, and accurately recognizing various oscillation modes is the prerequisite for understanding and controlling combustion instability. However, the high-dimensional spatial-temporal data of a complex combustion system typically poses considerable challenges to the dynamical mode recognition. Based on a two-layer bidirectional long short-term memory variational autoencoder (Bi-LSTM-VAE) dimensionality reduction model and a two-dimensional Wasserstein distance-based classifier (WDC), this study proposes a promising method (Bi-LSTM-VAE-WDC) for recognizing dynamical modes in oscillatory combustion systems. Specifically, the Bi-LSTM-VAE dimension reduction model was introduced to reduce the high-dimensional spatial-temporal data of the combustion system to a low-dimensional phase space; Gaussian kernel density estimates (GKDE) were computed based on the distribution of phase points in a grid; two-dimensional WD values were calculated from the GKDE maps to recognize the oscillation modes. The time-series data used in this study were obtained from numerical simulations of circular arrays of laminar flame oscillators. The results show that the novel Bi-LSTM-VAE method can produce a non-overlapping distribution of phase points, indicating an effective unsupervised mode recognition and classification. Furthermore, the present method exhibits a more prominent performance than VAE and PCA (principal component analysis) for distinguishing dynamical modes in complex flame systems, implying its potential in studying turbulent combustion.

[*] Corresponding author
  E-mail address: penzhang@cityu.edu.hk, Tel: (852)34429561.

[#] The authors equally contributed to the work.

# 1. Introduction

Combustion instabilities of annular combustion systems in aero engines and modern gas turbines have gained increasing attention. The complex combustion dynamics usually originate from the multi-scale space-time couplings of thermal-fluid systems [1]. Analyzing the dynamic characteristics directly from the high-dimension data of a complex combustion system typically presents significant challenges. In recent years, there has been a growing interest in reduced-order models (ROMs) of complex combustion phenomena [2-5]. The main idea is to reduce the dimensionality of the problem, capture the key features of combustion, facilitate the identification of combustion dynamics, and therefore enable the subsequent modulation and control of combustion instabilities.

In general, dimensionality reduction methods for modeling have two major categories: knowledge-based reduction and data-driven reduction. Although knowledge-based modeling is highly effective in extracting key physical information from relatively simple physical systems, it becomes increasingly difficult as the system complexity increases. Consequently, data-driven reduced order modeling offers an alternative approach to handling high dimensionality of complex problems. It typically has linear and non-linear approaches. Linear reduction approaches, including principal component analysis (PCA) [6], kernel principal component analysis (KPCA) [7], and proper orthogonal decomposition (POD) [8], are well-established and widely used in flow and combustion fields [9-11]. Their common prior assumption is that the original data can be represented by a linear combination of orthogonal basis functions. However, linear approaches often encounter difficulties in dealing with complex non-linear problems.

In recent years, machine learning has emerged as a potent methodology for investigating ROMs in flame systems [12-14]. The variational autoencoder (VAE) [15], a probabilistic variant of the autoencoder (AE), stands out for developing nonlinear ROMs for two reasons: its training is regularized to avoid overfitting; the latent space has good properties for the generative process [16]. Specifically, Omata and Shirayama [17] reported that the convolutional autoencoder is capable of reducing the dimensionality of data and representing the spatial-temporal structures of unsteady flows. Mrosek and Othmer [18] found that VAE outperforms POD for model order reduction in the vehicle flow fields. Arai et al. [19] enhanced VAE by incorporating an orthogonal decomposition layer, called the VAE-POD method, to study the nonlinear dynamics of combustion oscillations in rocket combustors. Iemura et al. [20] utilized VAE-POD to analyze the time series data of the cool-flame oscillation phenomenon. The combination of VAE and POD showed superiority over conventional POD and augmentation in the representation of input data, consequently aiding in understanding the underlying physical relationships among various variables. Previous studies have demonstrated that VAE has great potential in reducing data dimensionality and extracting meaningful features.

Many combustors employ multi-nozzle configurations with rotational symmetry due to the high energy density and combustion stability. However, the scenarios with breaking the symmetry are possible and common. As pointed out by Poinsot [1], breaking the symmetry of a combustor constitutes an appealing way to mitigate azimuthal instabilities. Yetter et al. [21] also proposed a promising new approach of asymmetric whirl combustion for non-premixed low-NOx combustor design. It should be noticed that Starship was designed to be powered by 33 engines firing in synchrony but ended with an explosion due to losing some engines on ascent [22]. Thus, the capacity to accurately monitor the combustion instability patterns in some extreme scenarios is important.

It is challenging to extract spatial-temporal information in circular flame systems with symmetry breaking, of which instability patterns exhibit pronounced temporal correlations. Learning temporal structure from a dataset, long short-term memory (LSTM) [23] emerges as a proven and effective architectural choice in sequential data analysis, prediction, and classification tasks in univariate and multivariate domains [24]. Recently, Zhu et al. used bidirectional-LSTM in the convolutional neural network to optimize the generation of time series sequences. Lei et al. [25] reported that the LSTM model has better accuracy and robustness for industrial temperature prediction.

Inspired by the above works, this study will integrate LSTM with VAE to study a complex nonlinear dynamical system consisting of identical flame oscillators in circular arrays with symmetry breaking. Utilizing the state-of-the-art technique can be beneficial to the method enrichment of nonlinear ROMs. Furthermore, the incorporation of the Wasserstein distance proposed in our previous work [26] in the low-dimensional latent space of VAE will be used to identify various dynamical modes of nonlinear flame oscillators with symmetric and asymmetric circular configurations.

## 2. Neural network model and numerical simulation

A two-layer bidirectional long short-term memory variational autoencoder (Bi-LSTM-VAE) model and a Wasserstein distance (WD) based classifier, called the LSTM-VAE-WD method henceforth, are for the first time designed to recognize and classify various synchronizing modes of circular nonlinear flame oscillators. Fig. 1 shows the detailed design of the present method, which is composed of three modules: two-layer Bi-LSTM-VAE, Bi-LSTM, and WD classifier. In addition, the datasets are obtained from numerical simulations of octuple identical flickering buoyant diffusion flames in a circular array to be expatiated shortly.

*2.1 Bidirectional long short-term memory variational autoencoder design*

The two-layer Bi-LSTM-VAE model mainly consists of two integral components: VAE and LSTM. VAE represents a variant of the AE extensively utilized as a potent dimensionality reduction tool in pattern analysis and machine intelligence fields. The fundamental structure of conventional VAE has three parts: an encoder, a reparameterization process, and a decoder. The encoder progressively reduces the dimensionality of the raw input data to a lower-dimensional latent space, while the decoder reconstructs the original data from the lower-dimensional space. The two coders consisting of multiple layers of fully connected neurons can be regarded as implicit nonlinear mapping functions $f$.

In the present study, the intricate dynamics of octuple flame oscillators in symmetric and asymmetric arrays are mapped into a distributed representation within a two-dimensional (2D) latent variable plane by using VAE in Fig.2. In the encoder, the correlation between the input $X = (x_1, x_2, x_3, \cdots, x_n)$, consisting of a set of variables $x_i$, and the output $Y = (\mu, \sigma^2)$, consisting of a two-variable vector, is briefly depicted by

$$Y = f(W_E X + b_E) \tag{1}$$

where $W_E$ and $b_E$ are the weight and bias of the encoder, respectively. To imbue the neural network with stochastic elements for improved gradient optimization, an external random disturbance following a Gaussian distribution of $\zeta \sim N(0, I)$, is introduced. Thus, the input of the encoder can be elegantly expressed as

$$Z = \mu + \sigma^2 \odot \zeta \tag{2}$$

which is commonly called the reparameterization trick; $\odot$ denotes the Hadamard product. Then the relationship between the decoder output $X' = (x_1', x_2', x_3', \cdots x_n', t')$ and the encoder input $Z = (z_1, z_2)$ is shown

$$X' = f(W_D Z + b_D) \tag{3}$$

where $W_D$ and $b_D$ are the weight and bias of the decoder, respectively. After the decoder, an additional layer of neurons is introduced for orthogonal decomposition. This layer is designed to transform the output of the decoder $X'$ through singular value decomposition into orthogonal outputs $X''$. This enhancement facilitates improved learning of the information necessary for pattern analysis within the network [20]. The relationship between the output of the decoder and the orthogonal decomposition layer is mathematically expressed as

$$X'' = (T\Sigma)^{-1}X' \tag{4}$$

where $T$ is the set of spatial eigenvectors and $\Sigma$ is the diagonal matrix of a singular value.

To acquire temporal information from data extracted from complex flame dynamics, we developed a two-layer Bi-LSTM network to substitute the fully connected layers of VAE, as illustrated in Fig. 3. In brief, LSTM preserves the temporal state of the network in hidden layer units referred to as memory cells, which are regulated by three kinds of gates: input gate, output gate, and forget gate. The input and output gates govern the input to the memory cell and the output flowing to other parts of the network, while the forget gate facilitates the transfer of output information from the previous time step to the next neuron. Information is stored in the memory cell by activating neurons through an activation function when the input unit exhibits high activation.

In this manner, the original input contains the time variable $t$, namely $X = (x_1, x_2, x_3, \cdots x_n, t)$. Specifically, $\{X_{t-k+1}, X_{t-k+2}, \cdots, X_t\}$ is the input data of the new encoder at the moment of $\{t-k+1, t-k+2, \cdots, t\}$, while $\{x_1, x_2, \cdots, x_n, t\}$ is the flame feature data. In the new designed decoder, $\{Z_{t-k+1}, Z_{t-k+2}, \cdots, Z_t\}$ denotes the input data at the corresponding moment, while $\{z_1, z_2\}$ is the two-feature data of the latent layer. Particularly, $\{h_1^i, h_2^i, \cdots, h_n^i\}$ denotes the output of the $i$th layer, where $i = 1$ and 2 for the two layers of Bi-LSTM.

The schematic diagram of the LSTM neuron is presented in Fig 4. The relationship between the input and output information of LSTM neurons is expressed by

$$f_t = sigmoid(W_{fg}X_t + W_{hfg}h_{t-1} + b_{fg}) \tag{5}$$

$$i_t = sigmoid(W_{ig}X_t + W_{hig}h_{t-1} + b_{ig}) \tag{6}$$

$$o_t = sigmoid(W_{og}X_t + W_{hog}h_{t-1} + b_{og}) \tag{7}$$

$$C_t = C_{t-1} \odot (f_t)_t + (i_t)_t \odot (\tanh(W_C X_t + W_{hC}h_{t-1} + b_C)) \tag{8}$$

$$h_t = o_t \odot \tanh(C_{t-1}) \tag{9}$$

where $W_{fg}$, $W_{hfg}$, $b_{fg}$ denote the weights and biases of the forget gate $f_t$, respectively. Similarly, $W_{ig}$, $W_{hig}$, and $b_{ig}$ are for the input gate $i_t$; $W_{og}$, $W_{hog}$, and $b_{og}$ are for output gate $o_t$; $W_C$, $W_{hC}$ and $b_C$ are for $C_t$. $t-1$ and $t$ are previous and current time steps. In the Bi-LSTM encoder and decoder, data is propagated in a bidirectional process of forward and backward paths.

LSTM possesses the capability to process historical information yet is unable to effectively learn future information. However, in the context of cyclic flame oscillation data, the information from both the preceding and succeeding moments demonstrates a strong temporal correlation with the current state. To address this limitation, this study employs a sophisticated two-layer Bi-LSTM network structure. Comprising two layers of bidirectional LSTM networks, each layer incorporates two distinct LSTM hidden layers, which yield similar outputs in opposing directions. For a given input sequence, the forward computation of the Bi-LSTM generates the output $h_t^{forward}$, while the backward computation generates the output $h_t^{backward}$. The final output of this layer is expressed as $h_t = [h_t^{forward}, h_t^{backward}]$.

In the present study, the two-layer Bi-LSTM-VAE model is trained through the minimization of mean squared error (MSE) and Kullback-Leibler (KL) divergence between the input and reconstructed data of flame time series. To train the neural network proposed in this study, we set four layers in the encoder and decoder. In the encoder, the first two are two-layer Bi-LSTM layers with 481 and 128 neurons respectively, followed by two fully connected layers with 256 and 2 neurons in the last two layers respectively. For the decoder, the four layers are two two-layer Bi-LSTM layers with 2–128 neurons and two fully connected layers with 256 and 481 neurons. The training strategy employs mini-batch training with a batch size of 64 and incorporates the early stopping mechanism. During the training, the neural network undergoes validation testing. To ensure optimal model preservation, model parameters are saved when a decrease in the validation set loss is observed. The Adam optimization algorithm is selected as the optimizer for the model. The training concludes either after 100 calculations of no improvement in validation set loss or upon reaching the pre-defined epoch limit of 2000. All code execution takes place about 9 hours for 18000 training samples on the NVIDIA Tesla V100SXM2 GPUs at the National Supercomputer Center in Guangzhou.

*2.2 Wasserstein distance-based classifier*

Inspired by Chi et al.'s recent systematical investigation of the dynamical modes of three flickering flames in the isosceles triangle arrangement [26], we adopt the Wasserstein-space-based methodology to classify various dynamical modes of the multiple flickering flames. The Wasserstein metric is widely used for time-series analysis [27-32]. Specifically, this study employs a two-layer LSTM-VAE model to project data from multi-flame systems onto a 2-dimensional phase space of two latent variables. The probability distribution of these latent space points is leveraged to identify distinct flame dynamical modes. To quantitatively assess the divergence in the distribution of different mode subsets within the latent space of multi-flame systems, the study adopts the 2-dimensional-Wasserstein distance (2D-WD) [33] to measure the closeness between two probability distributions in the latent space. For the same dynamical mode, their latent distributions are expected to be much closer than from other modes, consequently resulting in a smaller 2D-WD value.

As shown in Fig.5, the primary steps for computing 2D-WD in this study are outlined as follows:
(1) The first step is to normalize the two-dimensional distributions $Z_1$ and $Z_2$ and use the Gaussian kernel function to calculate the kernel density estimates. The calculation formula is shown in Eq. (10), where $\widehat{f_h}(z)$ represents the kernel density estimate at position $z$, $n$ is the sample size, $z_i$ is the observed value in the sample. $K_h$ is the Gaussian kernel function defined as Eq. (11), where $d$ is the dimensionality of the data, $h$ is the bandwidth, determined by Scott's Rule as Eq. (12), where the $cov$ is the covariance matrix of the data. Finally, we can obtain the Gaussian kernel density estimates for $Z_1$ and $Z_2$, denoted as $kde_1$ and $kde_2$, respectively.
(2) Then, the latent space is divided into a grid of $100 \times 100$ and point distributions of $kde_1$ and $kde_2$ in the meshed $Z_1$ and $Z_2$ plane are calculated for the probability density distribution. The two compared distributions are denoted as $P = (p_{ij})$ and $Q = (q_{ij})$, respectively.
(3) The last is to calculate the 2-dimensional Earth Mover's Distance between the two discrete probability distributions, as expressed in Eq. (13), where $f_{ij,kl}$ represents the flow from $p_{ij}$ to $q_{kl}$, $C_{ij,kl}$ is the Euclidean distance between elements $(i,j)$ and $(k,l)$. It is necessary to satisfy the conditions $\sum_{k,l} f_{ij,kl} = p_{ij}$, $\sum_{k,l} C_{ij,kl} = q_{ij}$, and $f_{ij,kl} \geq 0$, $C_{ij,kl} \geq 0$ simultaneously.

$$\widehat{f_h}(z) = \frac{1}{n}\sum_{i=1}^{n} K_h(z - z_i) \tag{10}$$

$$K_h(u) = \frac{1}{(2\pi)^{d/2} \cdot |h|^2} e^{-\frac{u^T u}{2h^2}} \tag{11}$$

$$h = n^{-1/(d+4)} \cdot cov^{1/2} \tag{12}$$

$$WD(P,Q) = EMD(P,Q) = minimize \sum_{i,j,k,l} f_{ij,kl} C_{ij,kl} \tag{13}$$

*2.3 Numerical simulations of flickering flames and data collection*

The vibratory motion of buoyant diffusion flames, first reported by Chamberlin and Rose as "the flicker of luminous flames" [34], has been studied for several decades [35-45]. It should be noted that Sitte and Doan [45] used the physics-informed neural network (PINN) to reconstruct unmeasured quantities from measured ones in a flickering pool fire. In recent years, it has been used as a nonlinear flame oscillator to investigate complex dynamics of multi-flame systems. Our previous works have computationally investigated flame oscillators in external swirling flows (single flickering jet flame) [46], 2-flame systems (dual flickering pool flames) [47], and 3-flame systems (triple flickering jet flames) [48]. These simulations have been validated quantitatively and qualitatively with experimental observations and successfully revealed the essential vortex-dynamical mechanisms of various dynamical modes in the flame oscillator systems.

Larger systems of flickering flames give rise to richer dynamical phenomena. Forrester [49] experimentally observed an initial-arch-bow-initial "worship" oscillation mode for four candles in a square arrangement. Manoj et al. [50] experimentally investigated the coupled behavior of annular networks with 5–7 candle-flame oscillators and observed variants of clustering and chimera states depending on the inter-flame distance and number of flames in the network.

As shown in Fig. 6(a), the present simulations of the circular flame system were carried out by arranging eight identical flickering flames at $Re$=100 along a circle with a diameter of $L = \alpha D$, where $D$ is the size of a single flame. The detailed descriptions of numerical schemes, parameters, and validations of present flame simulations are given in the Supporting Material, listed in Table S1, and shown in Fig. S1 respectively. The flame time series of three velocity components and temperature (4 physical variables) are collected from the 15 "sensors" distributed along the central axis of each flame. These sensors have an equal vertical distance of $D$ from the nozzle to the downstream, of which the range fully covers the dynamical behavior of each flame. In addition, this study simulated two groups of $L = 7.0D$ and $8.6D$, in which six different cases are carried out by changing the configuration symmetry.

As shown in Fig. 6(b), the six configurations are CA$\alpha$ case with octuple flickering flames: CA$\alpha$_m1 case with missing one flame at position 1, CA$\alpha$_m12 case with missing two flames at position 1 and 2, CA$\alpha$_m13 case with missing two flames at the position 1 and 3, CA$\alpha$_m14 case with missing two flames at the position 1 and 4, and CA$\alpha$_m15 case with missing two flames at the position 1 and 5. As the circular arrangement has rotational symmetry, the six configurations represent all scenarios, where a circular flame system possesses the symmetry of octuple flames and the asymmetry due to missing one or two flames. All sample data have the same 481 features (8 flames×15 sensors×4 physical variables+1 time variable) and cover 50 periods (5s) of single flickering flame with a sample frequency of 1000 Hz. The time and frequency domains of, for example, the velocity magnitude from the first sensors in CA7.0 and CA 8.6 circular flame systems show distinct modes with different amplitudes and frequencies (see more details in Fig. S2 of the Supporting Material).

**3. Results and discussion**

*3.1 Benchmark case: single flickering flame*

It is well known that the flame flicker is a self-exciting flow oscillation, which has been observed in diffusion, premixed, and partially premixed flames. Prominent experimental evidence was owing to Chen et al.'s [51] flow visualization of a methane jet diffusion flame, in which the large toroidal vortices outside the luminous flame are due to the buoyance-driven instability. Fig. 7(a) shows the temperature and vorticity of a single flickering flame (a nonlinear flame oscillator) during one period. As a benchmark, its 81-feature information along the flame center is collected to carry out the Bi-LSTM-VAE. It can be seen in Fig. 7(b) that there is a limit cycle repeating with time in the distribution of latent variables ($Z_1$, $Z_2$). The results show that the proposed Bi-LSTM-VAE method can effectively lower the high dimensional dynamics and map it into a low dimensional space while retaining the key physical information.

*3.2 Phase trajectories of a circular flame system*

Fig. 8 depicts the phase trajectories in the 2D latent space of Bi-LSTM-VAE and VAE as well as the two feature vectors of PCA corresponding to the first two maximum eigenvalues, where the phase points are colored with representing different dynamical modes. As outlined in Table 1, under the same constraint of two latent variables, the reconstruction error (mean square error, MSE) of Bi-LSTM-VAE output data from the original flame data is significantly lower than that of VAE and PCA, which indicates that Bi-LSTM-VAE outperforms VAE and PCA methods in reconstructing complex temporal-spatial information of flame system time-series data. The training and validating loss of the Bi-LSTM-VAE for the CA7.0 and CA8.6 groups can be seen in Fig. S3 of the Supporting Material.

Table 1 Reconstruction errors (MSE) of three approaches for the same flame time series data.

| Cases | Bi-LSTM-VAE | VAE | PCA |
|---|---|---|---|
| 7.0D | 0.150 | 0.242 | 0.563 |
| 8.6D | 0.185 | 0.298 | 0.566 |

By comparing the results of the three types of ROM methods, we can see that the original method of VAE and the linear method of PCA have mapped the high dimensional data into the more-or-less overlapped patterns in 2D space, while the proposed Bi-LSTM-VAE separates the latent space points of various flame modes without any intersection. This outstanding advantage can facilitate the mode recognition of various dynamical processes of complex systems, which are impossible to study in physical space due to the curse of dimensionality.

Furthermore, the lack of intersection among points from different modes is crucial to distinguish between dynamic patterns in the latent space. Otherwise, labeling the flame time series with a specific mode becomes challenging. In each group of CA7.0 and CA8.6, the latent-space distributions from different six scenarios are notably distinct, enabling the differentiation of dynamical modes. The detailed comparison of different modes in CA7.0 and CA8.6 groups can be seen in Fig. S4 of the Supporting Material. This latent space can be considered as a state space capable of distinguishing various oscillation modes of multi-flame systems.

*3.3 WD-based Classification*

To quantify the latent space distributions of Bi-LSTM-VAE, Fig. 9 plots the Gaussian kernel density estimation (GKDE) contours, which are derived from the latent points of ($Z_1$, $Z_2$). For the two groups of CA7.0 and CA8.6 in Fig. 9 (a) and Fig. 9 (b), the six benchmark mode data (left) and the corresponding test data (right) are illustrated. The color intensity in the plots signifies the magnitude of the GKDE values: higher values correspond to lighter colors, while lower values are represented by darker colors. By comparing the shape and the color distribution, we can see that the latent contour of each test data can be matched with the one benchmark for the six modes in the present study.

To further quantify the similarity between datasets, Fig. 10 presents the WD values between the benchmark and each of the six-test data for the CA7.0 and CA8.6 groups. Consistently, these values are amalgamated for both groups, with darker colors indicating smaller WD values, signifying a higher degree of similarity between the two distributions. The present results reveal that the diagonal elements exhibit the lightest colors. As shown in Fig. 9, the present WD-based classification aligns one-to-one with the actual dynamical modes, demonstrating the capability of the proposed Bi-LSTM-VAE-WD approach in mode recognition in complex multi-flame systems. The present results are consistent with our previous work on the mode recognition of three flame systems, where the dimension reduction of flames is carried out based on physical understanding.

**4. Conclusions**

An innovative deep neural network of bi-directional long short-term memory variational autoencoder (Bi-LSTM-VAE) and Wasserstein distance-based classifier (WDC) is proposed and used to identify dynamical modes of nonlinear flame oscillators in circular arrays with symmetry breaking. Besides, present numerical simulations of various dynamical modes of octuple flickering flame system systems have been carried out, where the buoyancy-driven flickering flame is solved by the unsteady, three-dimensional, low-Mach, and variable-density chemically reacting flows with a simplified chemical reaction mechanism. The dataset of flame time series consists of 481 features, of which four variables of $U$, $V$, $W$, and $T$ were sampled at 1000 Hz for 5s (about 50 flickering periods) along the central axis of each flame.

The established Bi-LSTM-VAE model is effective in mapping the spatiotemporal intricacies of the complex flame system into a low-dimensional space, where a two-dimensional WDC is for the first time integrated with the proposed neural network to classify dynamical modes from unlabeled data into the identified modes. The following conclusions are as follows. First, the comparison of the linear PCA, the conventional VAE, and the novel Bi-LSTM-VAE shows that the present approach outperforms the others in exhibiting non-overlapping point distributions in the latent space for various flame modes. This advantage serves as a distinguishable potential for the dimension reduction of complex combustion systems, for example, turbulent flames and real annular combustors. Second, the two-dimensional WDC method was applied for the quantitative assessment of the similarity between the distributions of two sets of flame time series. Accurate identification of flame dynamical modes is achieved by comparing their 2D-WD values and finding out the closest benchmark mode. The approach provides an unsupervised and effective approach for recognizing the dynamical modes of multiple flame systems.

We fully recognize that some challenging problems remain to be solved in future works, for example, the low-order modeling and bifurcations of various dynamical modes of large-scale flames, turbulent flame arrays, and circular-can turbine engine combustion, where the complex turbulence/chemistry interaction, the radiative heat loss, and the thermo-acoustic instability must be considered. The present work demonstrates the first successful attempt at solving these problems.

**Declaration of competing interest**

The authors declare that they have no conflict of interest.

**Acknowledgments**

This work is supported by the National Natural Science Foundation of China (No. 52176134) and partially by the APRC-CityU New Research Initiatives/Infrastructure Support from Central of City University of Hong Kong (No. 9610601).

**Supplementary material**

Supplementary material is available.

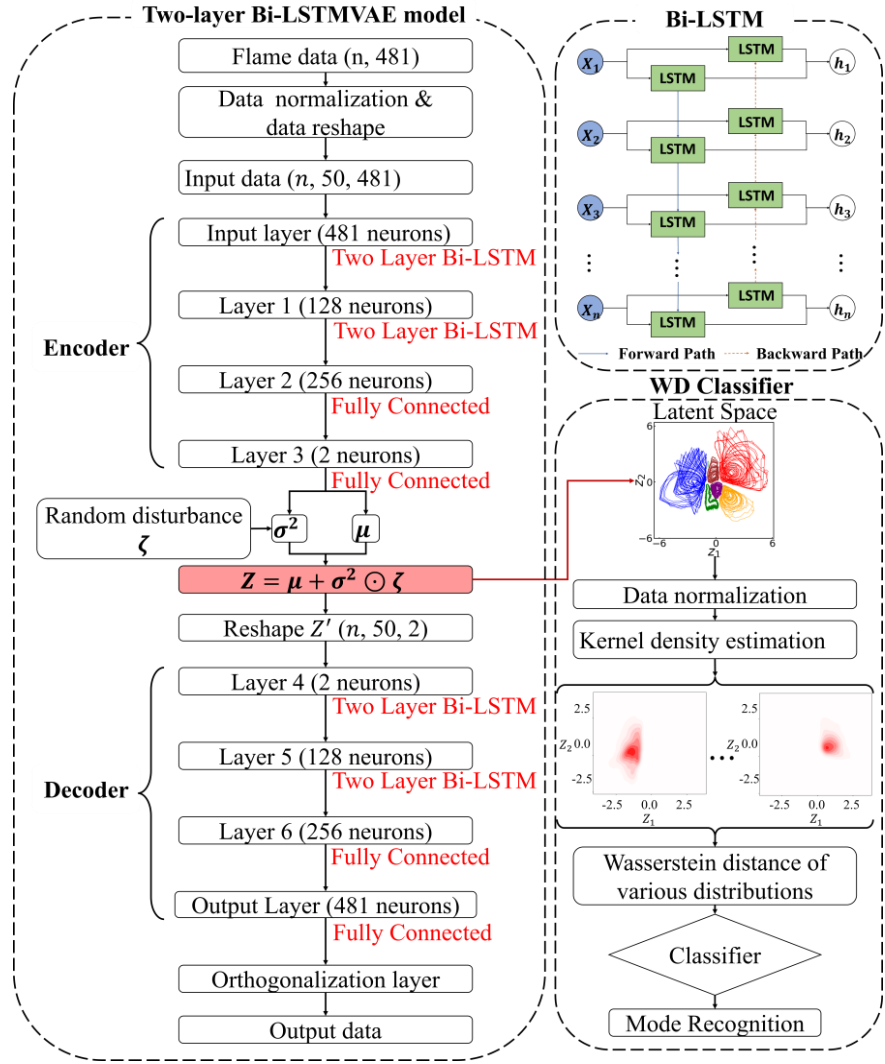

Fig. 1. Structure of the present Bi-LSTM-VAE-WD method consisting of three modules: a neural network structure of two-layer Bi-LSTM-VAE, a recurrent structure of Bi-LSTM, and a flowchart of WD classifier.

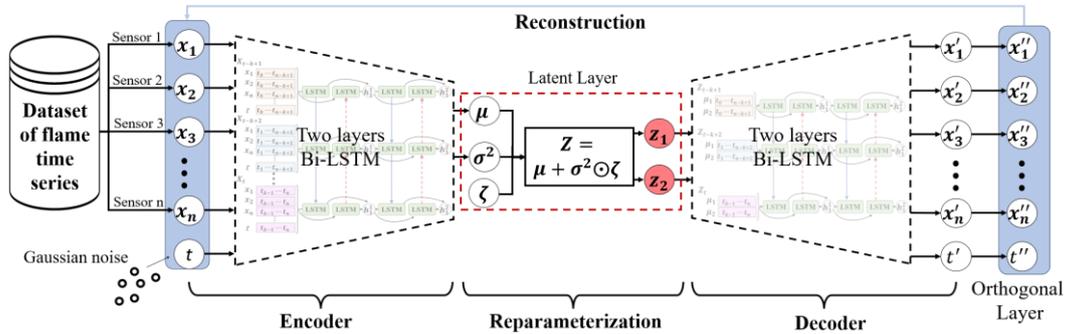

Fig. 2. Data flow through the present two-layer Bi-LSTM-VAE including the encoder and decoder of two-layer Bi-LSTM-VAE (The detailed Bi-LSTM structure in Fig. 3), the data flow through reparameterization, and the structure of orthogonal layer.

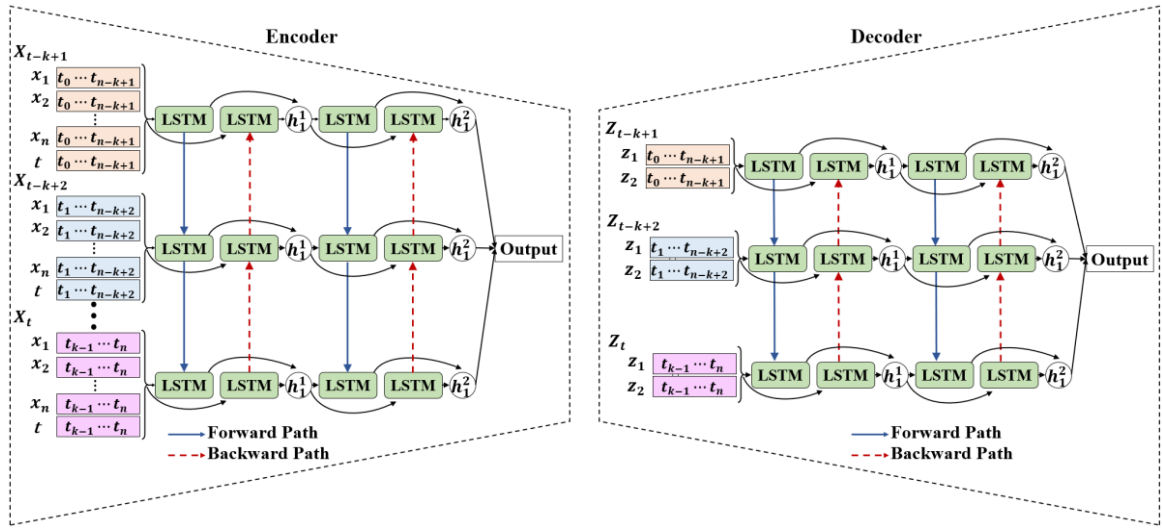

Fig. 3. Data flow of two-layer Bi-LSTM in Encoder and Decoder, each Bi-LSTM layer contains two computational paths in the forward and backward directions.

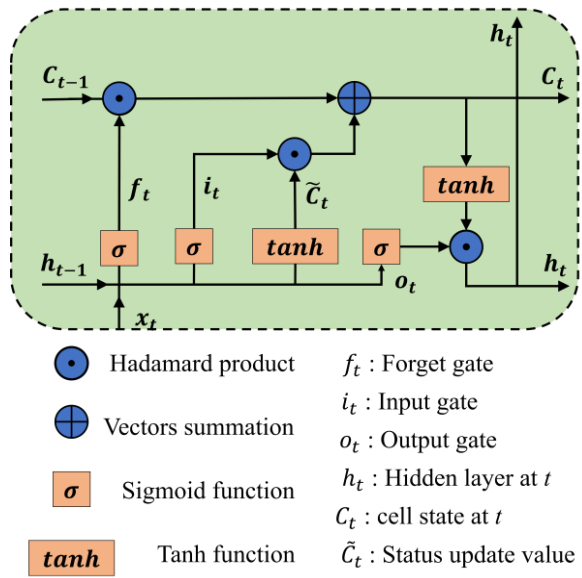

Fig. 4. Architecture of a single LSTM cell in the present two-layer Bi-LSTM.

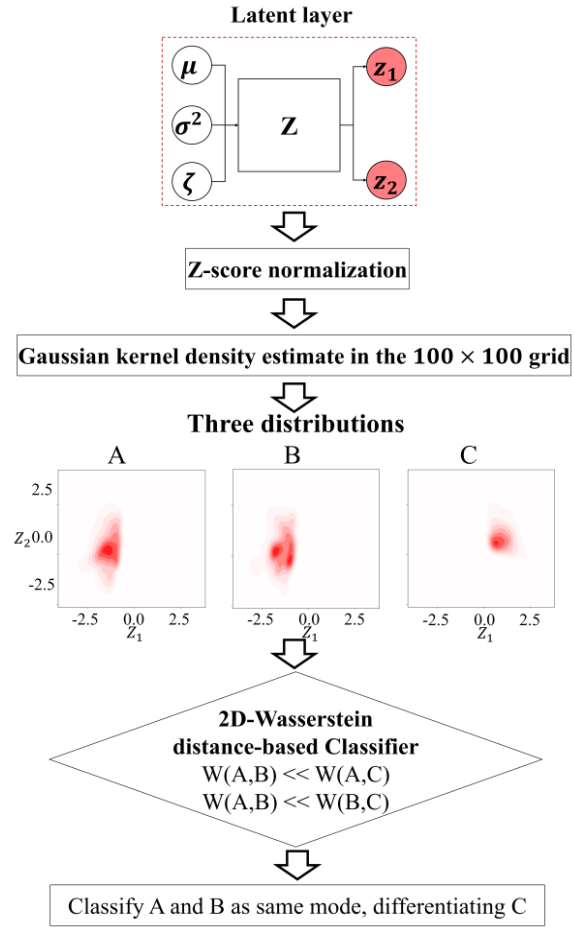

Fig. 5. Structure of the WD classifier. The structure consists of three parts: (1) Normalization of latent eigenvalue data. (2) Computing Gaussian kernel density estimates for latent variables. (3) Calculating 2D-Wasserstein-distance for distributions of benchmark and test data.

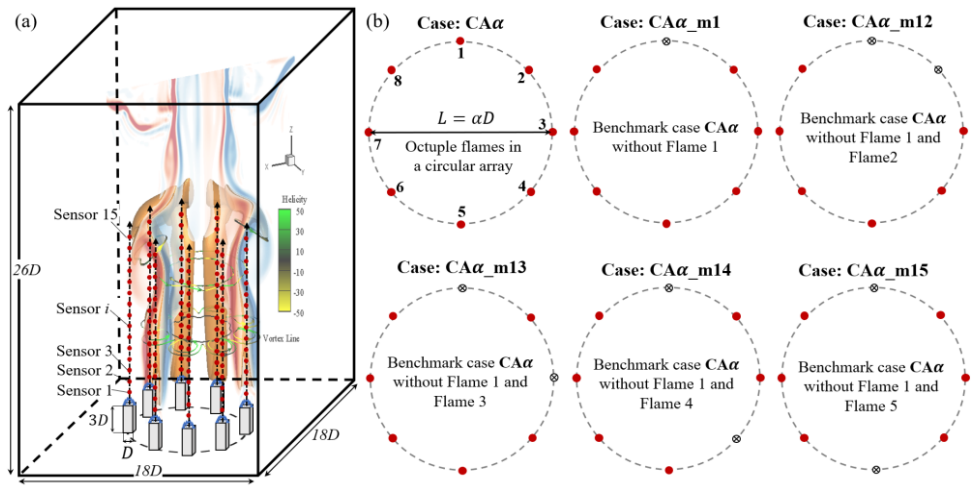

Fig. 6. (a) Computational setup of the octuple flame system and (b) the circular flame system with/without symmetry breaking.

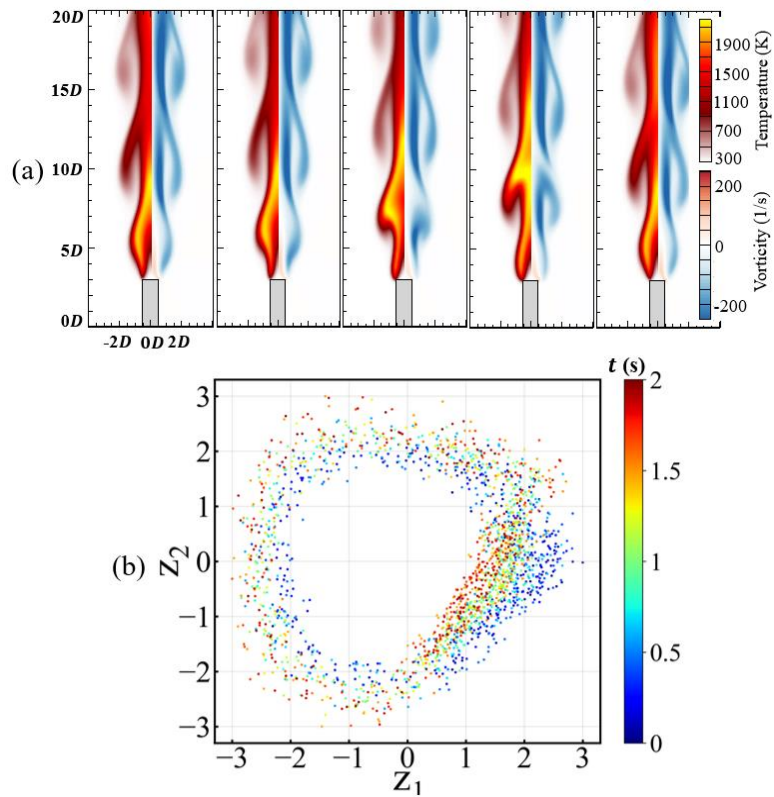

Fig. 7. (a) Physical feature of temperature (left) and vorticity (right) of single flickering buoyant diffusion flame. (b) Distribution of latent variables ($Z_1$, $Z_2$) of the nonlinear flame oscillator via Bi-LSTM-VAE. Data of velocity ($U, V$, and $W$) and temperature $T$ are collected from "sensors" along the central axis.

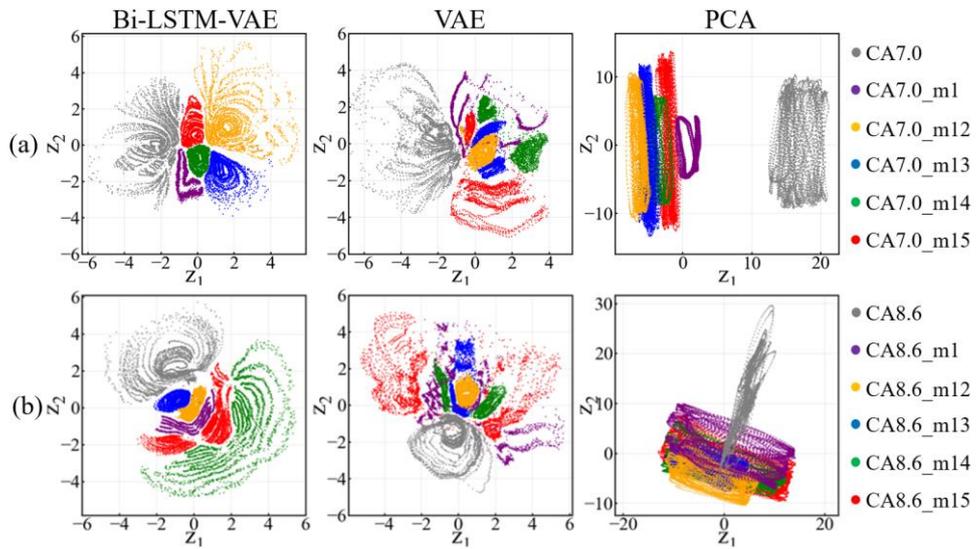

Fig. 8. Distribution of phase points of Bi-LSTM-VAE, VAE, and PCA for (a) CA7.0 and (b) CA8.6 groups.

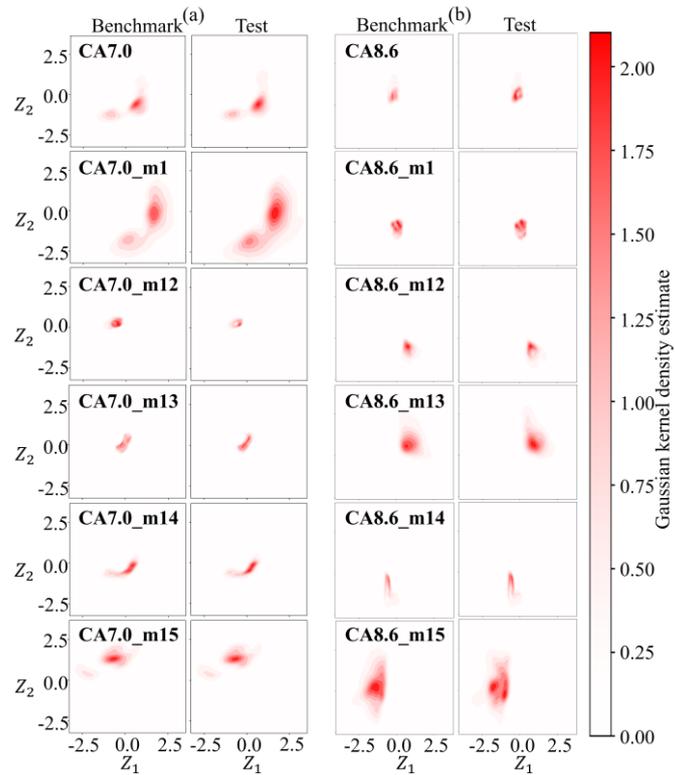

Fig. 9. Benchmark distributions of different dynamical modes are compared with those of the corresponding test data for (a) CA7.0 and (b) CA8.6 groups. Gaussian kernel density estimate is used to present the distribution of latent variables ($Z_1$, $Z_2$).

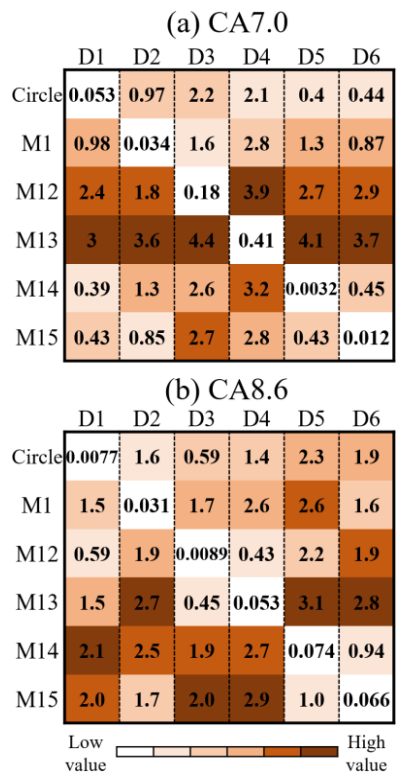

Fig. 10. Heat maps of WD calculated between benchmark cases (circle, M1, M12, M13, M14, and M15) and test cases (D1-6) for (a) CA7.0 and (b) CA8.6 groups.

# Supplementary Materials of Dimensionality Reduction and Dynamical Mode Recognition of Circular Arrays of Flame Oscillators Using Deep Neural Network

Weiming Xu[a,#], Tao Yang [b,#], and Peng Zhang[a,*]

*a Department of Mechanical Engineering, City University of Hong Kong, Kowloon Tong, Kowloon, 999077, Hong Kong*
*b Department of Mechanical Engineering, The Hong Kong Polytechnic University, Hung Hong, Kowloon, 999077, Hong Kong*

---

[*] Corresponding author

 E-mail address: penzhang@cityu.edu.hk, Tel: (852)34429561.

[#] The authors equally contributed to the work.

## 1. Numerical simulations of flickering flames in circular arrays

In the present study, flickering buoyant diffusion flames were computationally produced using Fire Dynamics Simulator (FDS) [1], which is a widely used open-source code for solving the unsteady, three-dimensional, low-Mach, and variable-density flow. The governing equations are used for computing the thermally driven flow:

$$\frac{\partial}{\partial t}(\rho) + \nabla \cdot (\rho \boldsymbol{u}) = 0 \tag{1}$$

$$\frac{\partial}{\partial t}(\rho Y_i) + \nabla \cdot (\rho Y_i \boldsymbol{u}) = \nabla \cdot (\rho D_i \nabla Y_i) + \dot{m}_i''' \tag{2}$$

$$\frac{\partial}{\partial t}(\rho \boldsymbol{u}) + \nabla \cdot (\rho \boldsymbol{u}\boldsymbol{u}) = -\nabla \tilde{p} - \nabla \cdot \sigma + (\rho - \rho_\infty)\boldsymbol{g} \tag{3}$$

$$\frac{\partial}{\partial t}(\rho h_s) + \nabla \cdot (\rho h_s \boldsymbol{u}) = \frac{D\bar{p}}{Dt} + \dot{q}''' - \nabla \cdot \dot{\boldsymbol{q}}'' \tag{4}$$

$$\rho = \frac{\bar{p}\overline{W}}{\mathcal{R}T} \tag{5}$$

where $\rho$ is the density, $\boldsymbol{u}$ the velocity vector, $Y_i$ and $D_i$ are the mass fraction and diffusion coefficient of species $i$ respectively, $\dot{m}_i'''$ is the mass production rate per unit volume of species $i$ by chemical reactions, $\tilde{p}$ is the pressure perturbation, $\sigma$ is the viscous stress, $\rho_\infty$ is the background density, $\boldsymbol{g}$ is the gravity vector, $h_s$ is the sensible enthalpy under low Mach number approximation, $\bar{p}$ is the back pressure, $\dot{q}'''$ is the heat release per unit volume, $\dot{\boldsymbol{q}}''$ is the heat flux vector, $\overline{W}$ is the molecular weight of the gas mixture, $\mathcal{R}$ is the universal gas constant, and $T$ is the temperature. Spatial integration was carried out using a kinetic-energy-conserving central difference scheme, and an explicit second-order predictor/corrector scheme advanced the time integration.

Based on FDS, our previous computational works [2-4] successfully captured the dynamical behaviors of flickering buoyant diffusion flames in a quiescent environment and reproduced a variety of dynamical modes in the dual- and triple-flame systems. More details of the simulation validations, including the scaling law of flickering frequency, the computing domain study, and the grid independent study, are given in [2, 3]. Fig. S1 shows the validation of flickering frequencies. The present simulations using infinite and finite fast one-step overall reaction agree fairly with previous experiments and particularly predict the famous scaling relation of $f_0 \sim \sqrt{gD}$. In addition, there is a very small difference on capturing the flickering frequency. Considering the computational cost, we use the mixing-limited chemical reaction to simulate the circular arrays of flame oscillators. Using a detailed reaction mechanism merits future study.

Table S1 Key parameters of numerical simulations

| | |
|---|---|
| Fuel | Methane gas with a uniform inlet velocity $U_0$=0.165 m/s |
| Bunsen burner | Eight identical square burners with $D$=10 mm length and 3D height |
| Domain | Whole computation in the $16D \times 16D \times 24D$ zone |
| Boundary | Six sides are open boundary condition for gas flowing in and out freely |
| Grid | $160 \times 160 \times 240$ structured, uniform, staggered grids. |
| Models | Mixing-limited chemical reaction (infinitely fast reaction of lumped species); no modeling for the insignificant influences of adiation and soot |
| All simulations | Dataset obtained from the fully developed flows after running all cases at least 20 times longer than $L/U_0$) |

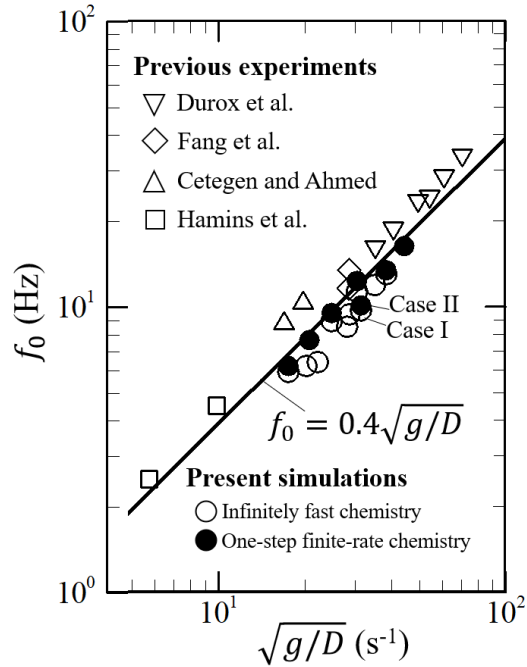

Fig. S1. Validation of single flickering flames with the scaling laws [5] and previous experiments [6-9]. The numerical results are obtained by using infinitely fast chemistry (IFC) and one-step finite rate chemistry (OFC).

## 2. Time and frequency domain information of circular arrays of flame oscillators

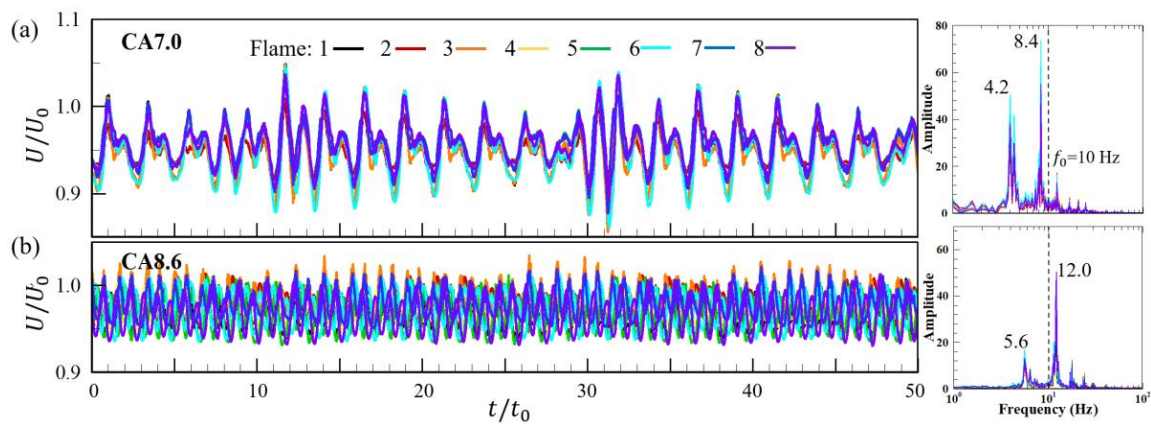

Fig. S2. Time and frequency domains of, for example, the velocity magnitude extracted from the first "sensor" in the two cases of (a) CA7.0 and (b) CA8.6 without missing any flame.

## 3. Phase trajectory of various flame modes

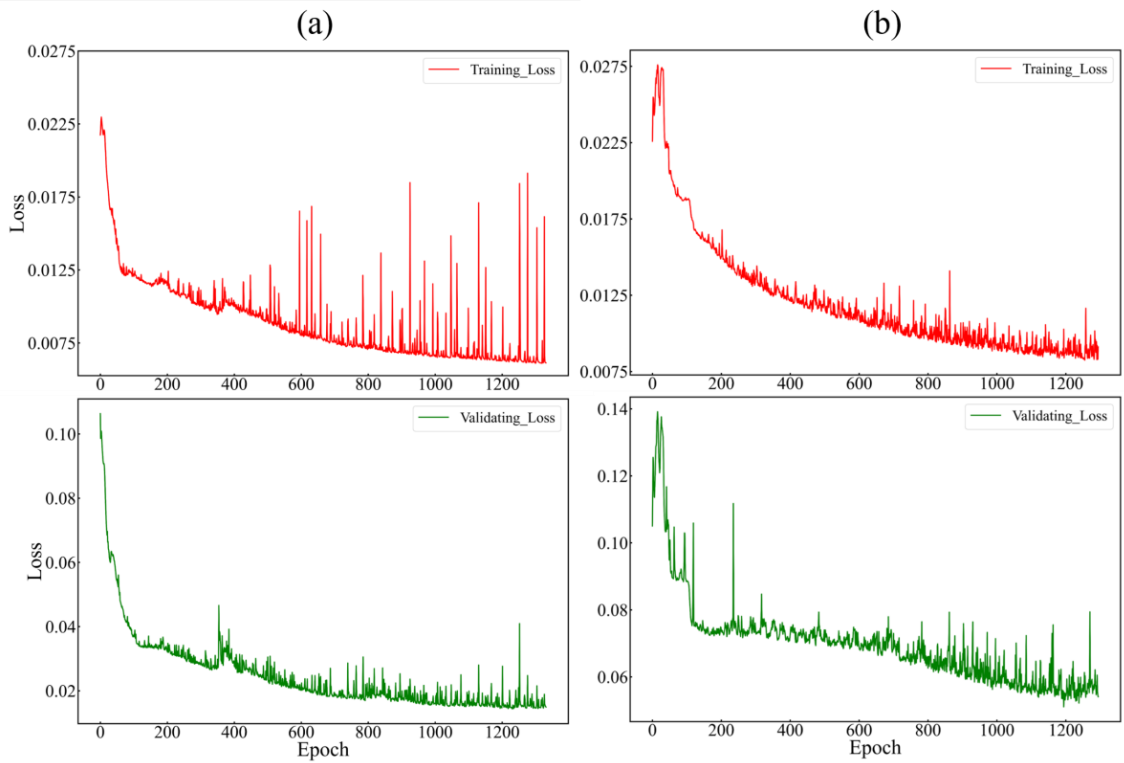

Fig. S3. Training and validating loss of two cases (a) CA7.0 and (b) CA8.6.

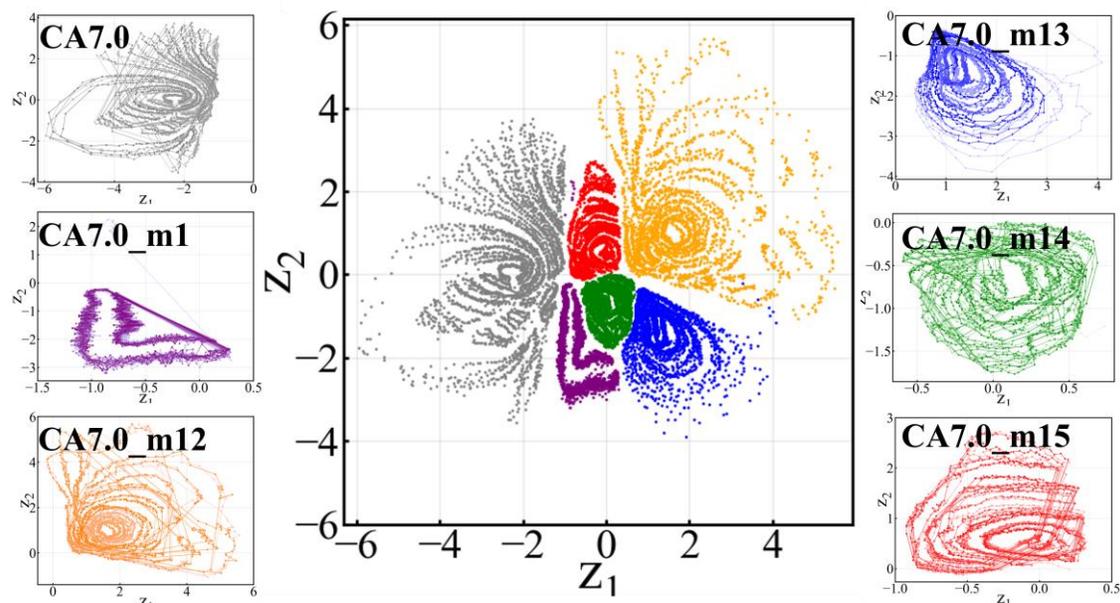

(a) CA7.0 group

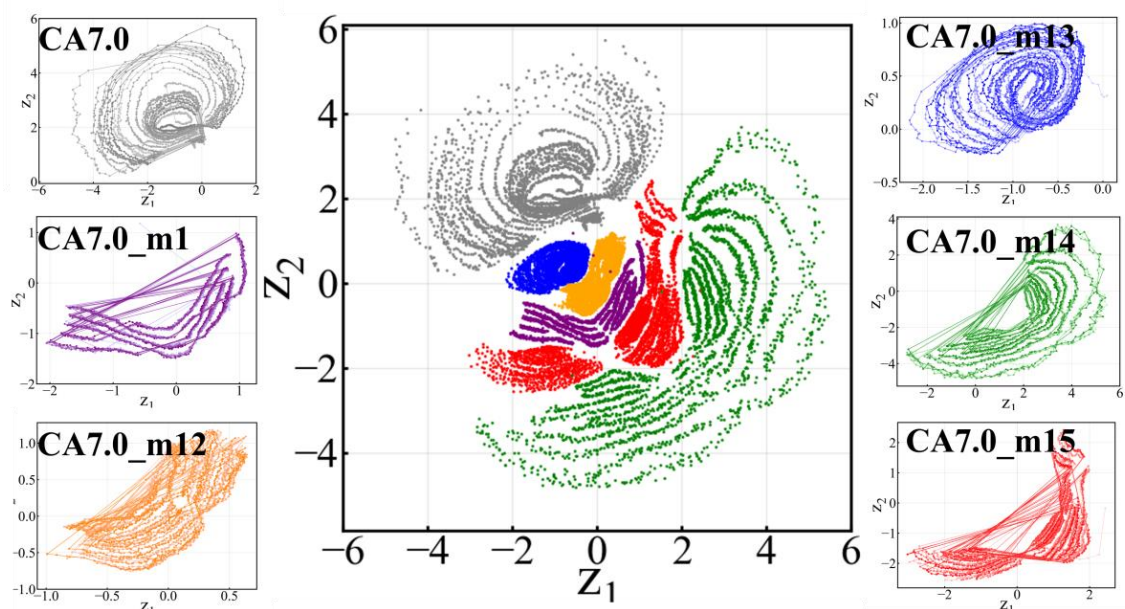

(b) CA8.6 group

Fig. S4. Latent Plane of Bi-LSTM-VAE for six flame modes in (a) CA7.0 and (b) CA8.6 groups.

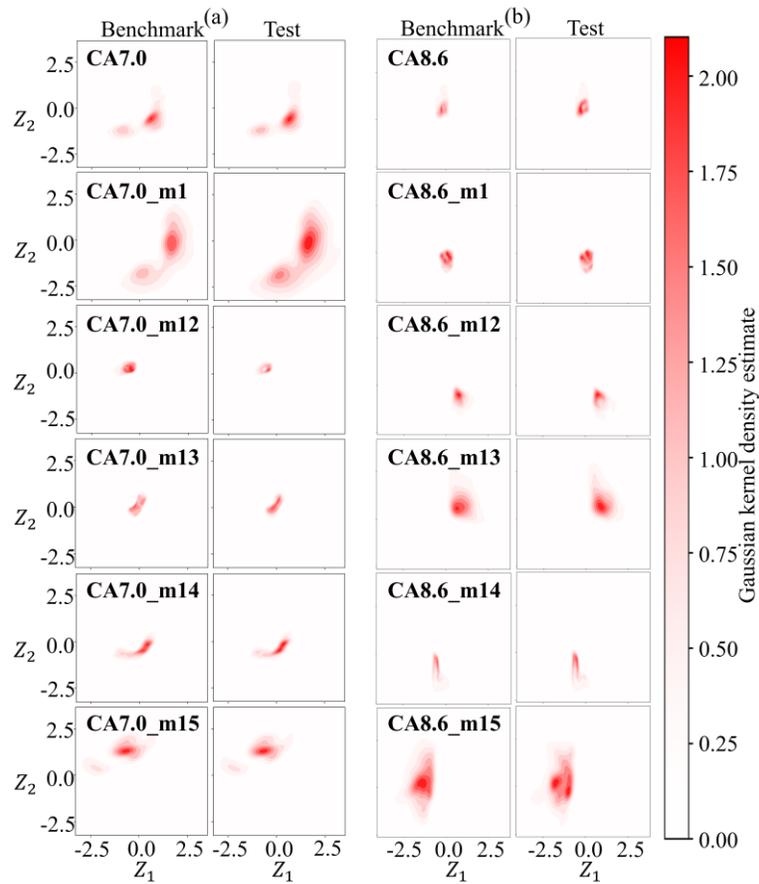

Fig.S5 Benchmark distributions of different dynamical modes are compared with those of the corresponding test data for (a) CA7.0 and (b) CA8.6 groups. Gaussian kernel density estimate is used to present the distribution of latent variables ($Z_1$, $Z_2$).